\documentclass[10pt,twocolumn,letterpaper]{article}

\usepackage{cvpr}
\usepackage{times}
\usepackage{epsfig}
\usepackage{graphicx}
\usepackage{amsmath}
\usepackage{amssymb}
\usepackage{multirow}
\usepackage{array}
\usepackage{booktabs}
\usepackage{colortbl}
\usepackage{makecell}
\usepackage{bbding}
\usepackage{eso-pic}

\usepackage[breaklinks=true,bookmarks=false]{hyperref}

\cvprfinalcopy 

\def\cvprPaperID{} 

\ifcvprfinal\fi
\begin{document}
	
	\title{3D-SSD: Learning Hierarchical Features from RGB-D Images for Amodal 3D Object Detection}
	
	\author{Qianhui Luo, Huifang Ma, Yue Wang, Li Tang, Rong Xiong\\
		Institute of Cyber-Systems and Control, Zhejiang University\\
		{\tt\small $\{$qianhuiluo, hfma, rxiong$\}$@zju.edu.cn, wangyue@iipc.zju.edu.cn, litang.cv@gmail.com}
		\and
	}
	
	\maketitle
	
	\begin{abstract}
		This paper aims at developing a faster and a more accurate solution to the amodal 3D object detection problem for indoor scenes. It is achieved through a novel neural network that takes a pair of RGB-D images as the input and delivers oriented 3D bounding boxes as the output. The network, named 3D-SSD, composed of two parts: hierarchical feature fusion and multi-layer prediction.
		The hierarchical feature fusion combines appearance and geometric features from RGB-D images while the multi-layer prediction utilizes multi-scale features for object detection. As a result, the network can exploit 2.5D representations in a synergetic way to improve the accuracy and efficiency. The issue of object sizes is addressed by attaching a set of 3D anchor boxes with varying sizes to every location of the prediction layers. At the end stage, the category scores for 3D anchor boxes are generated with adjusted positions, sizes and orientations respectively, leading to the final detections using non-maximum suppression. In the training phase, the positive samples are identified with the aid of 2D ground truth to avoid the noisy estimation of depth from raw data, which guide to a better converged model.
		Experiments performed on the challenging SUN RGB-D dataset show that our algorithm outperforms the state-of-the-art Deep Sliding Shape by 10.2\% mAP and 88$\times$ faster. Further, experiments also suggest our approach achieves comparable accuracy and is 386$\times$ faster than the state-of-art method on the NYUv2 dataset even with a smaller input image size.
	\end{abstract}
	\section{Introduction}
	\begin{figure*}[t]
		\centering
		\includegraphics[width=1\textwidth]{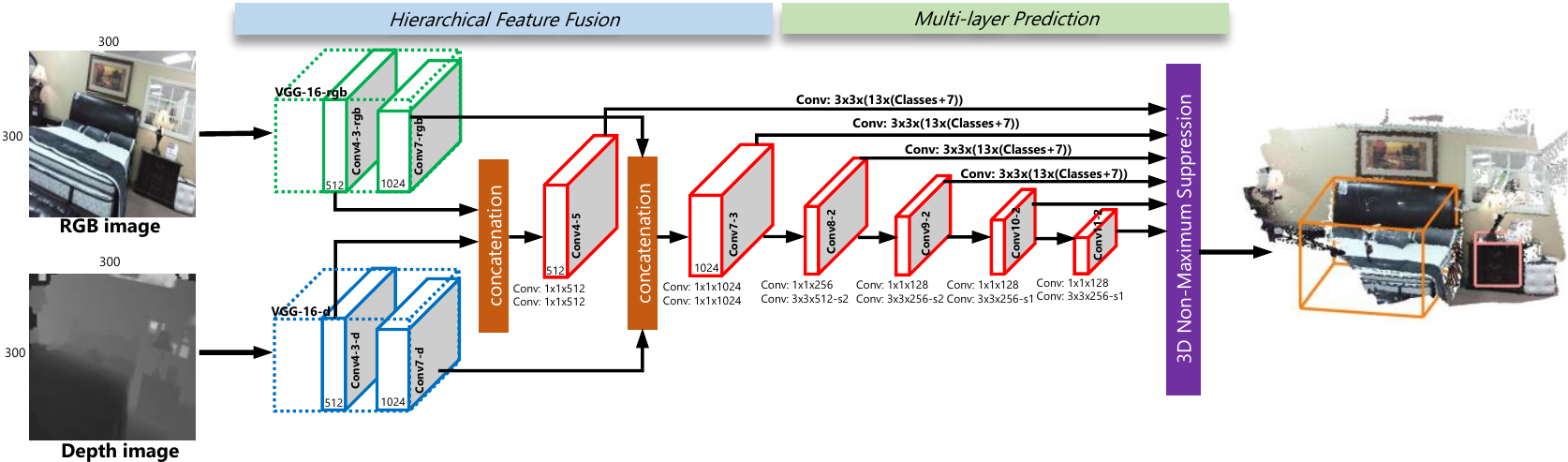}
		\caption{{\small \textbf{3D-SSD Architecture.} The network takes a pair of  RGB-D images as the input and hierarchically fuses appearance and geometric features from the RGB and depth images. Then, a set of 3D anchor boxes with manually-set sizes are attached to every location of the prediction layers, whose 3D positions are estimated from the depth image. Without using 2D bounding box proposals, these multi-scale prediction layers are used to estimate category scores for each anchor box and regress its position, size and orientation offsets.
		}}
		\label{model}
		\vspace{-3mm}
	\end{figure*}
	
	Object detection has been a popular area of research with reported high accuracies in fast computational times mainly due to the presence of widely available large-scale datasets\cite{VOC2010,COCO2014,imagenet2015} and the developments in convolution neural networks(ConvNets). The aim has been to produce 2D bounding boxes around objects on the image plane along with their identified categories. However, real world applications such as robotic object handling, mapping and augmented reality demand 3D information such as pose, size and geometric position. 
	
	
	As a result, amodal 3D object detection in indoor scenes has gained growing attention especially after the launch of cheaper RGB-D sensors (\eg Kinect and Xtion). The task has been to produce a 3D bounding box surrounding the full extent of a object in the real world even under partial observations. Current solutions for this problem can be broadly divided into two categories. The first category, \emph{2.5D approaches} \cite{K2013accurate, Wei2015DBM,Saurabh2015depthrcnn,Deng2017}, encodes depth as an extra channel of the color images in order to make full use of the successful 2D ConvNets followed by association of 2D proposals to the counterpart 3D boxes. The second category, \emph{3D approaches}, converts the depth image into a point cloud followed by design of 3D features\cite{Surans2014sliding, Zhile2016COG} or develop 3D ConvNets\cite{Suran2016DSS} to better explore geometric information of the 3D representation.\par
	Although there are convincing and inspiring recent works in both directions, further work is mandatory due to some unresolved issues: (1)  The proposal generation from RGB-D images is computational intensive. As it is indispensable in prediction phase, the network is thus inapplicable for real-time demands. (2) Different to 2D two-stage detector, most popular proposal generators in 3D are non-differentiable or independent to later recognition network, thus cannot be optimized together, losing the end-to-end performance. (3) Features of sparse 3D data are hardly distinctive. When projecting pixels of image back to 3D space, the generated 3D point clouds are usually noisy and sparse due to the low resolution and the perspective occlusion, constraining the detections to large objects with salient shape features.\par
	
	In view of the above problems, we set to build an end-to-end RGB-D object detector in the third way, which is to predict the dense bounding boxes. The proposed network termed \textbf{3D-SSD}, takes a pair of RGB-D images as inputs and predicts the full extent of objects in 3D space. As illustrated in Figure \ref{model}, after the prior feature extraction network, the 3D-SSD consists of two parts: \emph{hierarchical feature fusion} and \emph{multi-layer prediction}. In the \emph{hierarchical feature fusion} part, two subnetworks are used to learn appearance and geometric features from the RGB and depth images respectively, and then the features are fused on multiple stages of the network. The \emph{multi-layer prediction} part is a 3D generalization of the popular SSD \cite{Wei2016SSD} framework. 
	To address the issue of the scale ambiguity, we attach a small number of 3D anchor boxes with different sizes to each location of the prediction layers. The depth image is utilized to determine the 3D poses of these anchor boxes. Finally, small convolutional filters are applied to these multi-scale feature layers to refine the position, size and orientation offsets relative to each anchor box followed by predicting its object category. \par
	
	We evaluate our method on the challenging SUN RGB-D and NYUv2 datasets. Experiments show the 3D-SSD outperforms the state-of-the-art \emph{Deep Sliding Shape}\cite{Suran2016DSS} \cite{Deng2017} in both accuracy and efficiency. Our contributions can be summarized as follows:\par 
	(1) A novel end-to-end framework (3D-SSD) for amodal 3D object detection is proposed. It takes a pair of RGB-D images as the input and it accurately predicts positions, sizes, orientations and categories of objects in 3D space at a much faster speed.\par
	
	(2) The proposed hierarchical fusion structure incorporates the appearance and geometric features from RGB and depth images comprehensively by concatenating feature layers on different stages of the network. This gives consideration to both of the fine details  and high-level semantics of an object. \par
	
	(3) A 2D bounding box ground truth aided matching strategy is proposed to identify 3D positive samples during the training stage as for the sake of robustness. \par
	(4) Experimental results have confirmed the effectiveness of utilizing multi-scale 2.5D information for amodal 3D object detection.\par

	\section{Related works}
	Object detection has progressed rapidly since the seminal work of RCNN \cite{R2014rcnn} used ConvNets to predict a bounding box around the visible part of an object on the image plane. This work attracted much attention and followed by Fast-RCNN\cite{R2015Fastrcnn}, Faster-RCNN\cite{He2014fasterrcnn}, YOLO\cite{J2016YOLO, J2017YOLO9000}, SSD\cite{Wei2016SSD}, Mask-RCNN\cite{He2017maskrcnn},  Focal Loss\cite{He2017focal} to gradually improve the accuracy and speed. All the above results are in 2D and there were some earlier works on 3D object detection starting to emerge \cite{K2013accurate, D2013Holistic}. The work given in \cite{K2013accurate} produces multiple hypothetical object foreground masks for each detected 2D bounding box,  and then uses corresponding  point cloud to extract handcrafted features for each mask. The DPM algorithm\cite{DPM2010} generalized to RGB-D images is used to determine the 3D position for the foreground object. The work given in \cite{D2013Holistic} firstly  applies  the CPMC algorithm\cite{CPMC2012} to generate candidate cuboids.  Features extracted from  RGB-D images and  contextual relations are used to assign semantic labels for these cuboids. Then the success of ConvNets and the arrival of 3D sensors began to drive a new era in 3D detection. In the following paragraphs, We briefly review existing works on 3D object detection with RGB-D images, along with the topic of prediction using multiple layers and multi-feature fusion methods.\par
	\vspace{1mm}
	\noindent\textbf{3D object detection in RGB-D images:} \emph{Sliding shapes} \cite{Surans2014sliding} converts a depth image into a point cloud and then slides 3D detection windows in the 3D space. Handcrafted features of each 3D window are later fed into exemplar SVMs for each object class. An exemplar SVM is trained by rendering a CAD model into synthetic depth images. Repeatedly computing 3D features of each sliding window in the 3D space followed by applying many exemplar SVMs cause this method to perform very slow. Similar to \cite{Surans2014sliding}, \cite{Wei2015DBM} also adopts a sliding-window fashion in the 3D point cloud, where raw features for each 3D box are computed by applying two RCNNs to its 2D projection on the RGB and depth images separately. Then a deep Boltzmann Machine trained with CAD models is utilized on raw features to exploit cross-modality features from RGB-D images. In the last stage, exemplar SVMs are used for object detections. In the same year, \cite{Saurabh2015depthrcnn} employed the RCNN to estimate a coarse pose for each instance segmentation in the 2D image by encoding a depth image as an input 3-channels image (local surface normal vector $N_{x}, N_{y}$ and $N_{z}$ w.r.t gravity estimation), and then aligned the CAD model to best fit the point cloud inside the segmentation using ICP. \par
	
	Inspired by the Faster-RCNN, \emph{Deep Sliding Shape} \cite{Suran2016DSS} divides a 3D scene recovered from RGB-D images into 3D voxel grids, and designs a 3D ConvNet(named 3D RPN) to extract 3D region proposals to better explore 3D representations. Then, for each 3D proposal, another 3D ConvNet(named 3D ORN) is used to learn object categories and regress 3D boxes. The use of 3D ConvNets significantly outperforms \cite{Surans2014sliding} both on the accuracy and the efficiency. The work by \cite{Zhile2016COG} designs clouds of orientated gradients for a sliding 3D box leading to dramatical improvement of the performance \cite{Surans2014sliding}.\par
	
	To reduce the search space in 3D, \cite{Jeab20172DD} makes use of the Faster-RCNN to detect objects in the RGB image and then uses the point cloud inside each 2D bounding box to regress positions and sizes in 3D by using a multilayer perceptron network. Among all the aforementioned works, \cite{Surans2014sliding}, \cite{Suran2016DSS}, \cite{Zhile2016COG} and \cite{Jeab20172DD} have made Manhattan world assumption and sliding 3D boxes were aligned with the estimated room layout. The work given in \cite{Deng2017} chooses time-consuming externed multiscale combinatorial grouping(MCG) \cite{P2014MCG} in RGB-D images \cite{S2014rich} to obtain 2D bounding box proposals. Based on the Fast-RCNN, features of each 2D proposal on RGB and depth images were integrated for classification and 3D box regression. \par
	
	In this paper, we do not restructure point cloud from depth images since it is usually noisy and  time-consuming. Our method is a single end-to-end framework which operates in 2D images and does not requre 2D bounding box proposals.\par
	\vspace{1mm}
	\noindent\textbf{Prediction using multiple layers:} A number of recent approaches use multi-scale feature layers in a ConvNet to improve 2D object detection (\eg SSD \cite{Wei2016SSD} and FPN \cite{T2017FCN}). While they achieve inspiring performances, none of the exiting works have utilized multiple layers in ConvNets for 3D object detection in RGB-D images. In this work, we adopt multi-scale feature maps to improve the performance of 3D object detection.\par 
	\vspace{1mm}
	\noindent\textbf{Multi-feature fusion:} In most existing works, features are combined from RGB and depth images in the late stage of a deep neural network for 3D object detection. Wei \etal \cite{Wei2015DBM} uses a deep Boltzmann Machine to fuse features from two independent RCNNs. The works in \cite{Suran2016DSS}, \cite{S2014rich}, \cite{Deng2017} concatenate the features from RGB-D images directly and \cite{Multiview2017} designs a deep fusion network to combine region-wise features from multiple views before predictions. Our network differs from previous methods because we hierarchically fuse features on both earlier and middle stages.\par

	\section{3D-SSD Network}
	
	The 3D-SSD network takes a pair of RGB-D images as input and hierarchically fuses appearance and geometric features from the RGB and depth images. Then, a set of 3D anchor boxes with manually-set sizes are attached to every location of the prediction layers, whose 3D positions are estimated from the depth image. Without using 2D bounding box proposals, these multi-scale prediction layers are used to estimate category scores for each anchor box and regress its position, size and orientation offsets.
	
	\subsection{Hierarchical Feature Fusion}
	
	In the prior literature, researchers have used \emph{early fusion} \cite{Cai2016early} or \emph{late fusion}\cite{Suran2016DSS} \cite{Deng2017} to combine features from different input data. Early fusion firstly concatenates features from different images and then feeds the fused features into a neural network. Late fusion uses independent subnetworks to process multi-inputs and then concatenates their feature maps to predict detections. In order to integrate the details captured on the early stage of a network with the high-level semantics modeled on the late stage, we propose a hierarchical fusion structure, which incorporates features from different inputs on both stages.\par
	
	The input RGB and depth images (all resized to $300\times300$) separately go through two VGG-16 networks to learn appearance and geometric features, as shown in Figure \ref{model}. Then, we have selected two pairs of layers with identical receptive field in the network to fuse their features. 
	Specifically, in the \emph{hierarchical feature fusion} part, we firstly concatenate the convolution layers of conv4-3-rgb and conv4-3-d, which are later followed by two $1\times1$ convolution layers to shuffle and choose appearance and geometric features that come from the same region on the original input RGB and depth images. Then is the concatenation of convolution layers conv7-rgb and conv7-d, also followed by two $1\times1$ convolution layers. The fused feature maps on these two stages are parts of the multiple feature layers used for prediction in the \emph{multi-layer prediction} part. \par
	
	\begin{figure}[ht]
		\centering
		\includegraphics[width=0.35\textwidth]{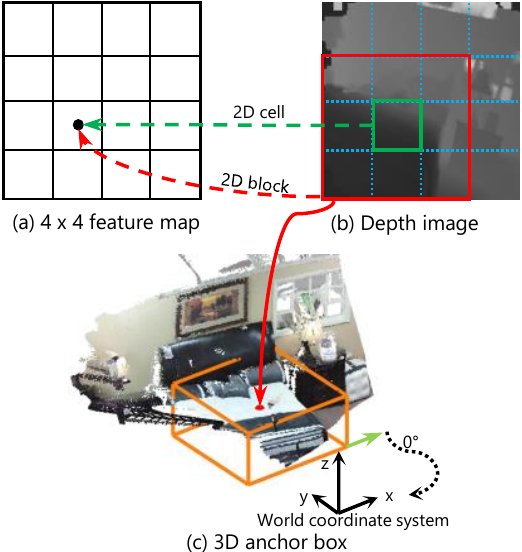}
		\caption{\textbf{An example of generation for 3D anchor boxes.} (a) shows a exemplar feature map($4\times4$) in our model, the black dot represents its location in the $3^{rd}$ row and $2^{nd}$ column. (b) The green and red squares are an illustration for 2D cell and 2D block belong to the $(3,2)$ location in (a). The red dot in (c) shows the common centers of 13 anchor boxes calculated from 2D block. The orange cuboid is one 3D anchor box out of 13. The orientation angle of its longer edge(cyan vector) in $xy$-plane to $x$-axis is $0^{\circ}$.}
		\label{anchor}
		\vspace{-3mm}
	\end{figure}    
	\subsection{Multi-layer Prediction}
	The 2D features from images cannot be directly used for 3D object detection without other calibration information due to the scaling problem. 
	Without using 2D bounding box proposals, we apply a small(3$\times$3) convolutional filter(ConvFilter) on every location of the prediction layers to detect objects. When the small ConvFilter acts on the lower level feature maps, its receptive field is limited and can only capture local features of  some of the big objects. While upper level feature maps possessed with high-level semantics may miss some fine details of the small objects. To address this issue, we follow the prediction paradigm of the SSD \cite{Wei2016SSD} by incorporating multi-scale feature layers to improve performance of 3D object detection. As is shown in Figure \ref{model}, six convolution layers conv4-5, conv7-3,  conv8-2, conv9-2, conv10-2 and conv11-2 in the \emph{multi-layer prediction} part are used to produce a number of 3D bounding boxes, which are later fed to a 3D non-maximum suppression to estimate the final object detection results. \par
	
	\vspace{1mm}
	\noindent\textbf{3D anchor boxes:} We have attached a set of center-aligned 3D anchor boxes to each location of the prediction layers to indicate the possible object candidates. Since the visible parts of an object on images may not reflect their real size due to different view points from cameras, lower feature layers with small receptive field can also fit for some big objects in 3D space and vise versa for upper feature layers. We have fixed the sizes for the 3D anchor boxes attached to each location, which are set to be 13 kinds based on the statistics of ground truth object sizes on the training set.\par 
	Inspired by \cite{Deng2017} that 3D proposals can be initialized from 2D segment proposals, we propose to estimate the 3D position of these anchor boxes from the corresponding \textbf{2D blocks} on the depth image, as illustrated in Figure \ref{anchor}.
	For a feature layer of size $m$$\times$$n$ with $p$ channels, at each of the $m$$\times$$n$ locations, we define a square region on the initial input image as its \textbf{2D cell} by dividing the input image into $m$$\times$$n$ grids. And the grid in $i^{th}$ row and $j^{th}$ column is the 2D cell related to the location in $i^{th}$ row and $j^{th}$ column of the feature layer. Then its \textit{2D block} is composed of its 2D cell along with the $8$ neighboring cells, namely a region of $3$$\times$$3$ cells.
	Since the depth images are usually noisy and sparse, we use the median value $z_{med}$ of the 2D block to approximate the actual depth value of the 2D block center. \par
	Then the 3D position $(x_{0},y_{0},z_{0})$ of anchor boxes can be obtained by projecting the approximated 2D block center $(c_{x},c_{y},z_{med})$ in the camera coordinate to the world coordinate(see Figure\ref{anchor}) using camera intrinsic and extrinsic parameters:\par 
	\begin{equation}
	\vspace{-1mm}
	\begin{pmatrix} x_{0} \\ y_{0} \\ z_{0} \end{pmatrix} = R_{tilt} * \begin{pmatrix} z_{med} * (c_{x} - o_{x}) / f_{x} \\ z_{med} * (c_{y} - o_{y}) / f_{y} \\ z_{med}\end{pmatrix}
	\end{equation}
	where $(o_{x}, o_{y})$ is the principal point, $(f_{x},f_{y})$ is the focal length of camera, $R_{tilt}$ is the transform matrix between camera and world system. \par
	
	We denote a 3D bounding box by its center position $(x_{0}, y_{0}, z_{0})$ and  size $(w, l, h)$ in the world coordinate system, along with its orientation angle $\theta$, defined as its rotation around the $z$ axis. In the initial phase of the experiment, the orientation angle $\theta$ for each anchor box is set to be $0^{\circ}$, that is, the edges of each anchor box are separately parallel with the three axises of world system. For each 3D anchor box, the small ConvFilter applied to this location then computes regression scores for the total $c$ target classes and 7 offsets relative to its position, size and orientation. Thus, the ConvFilter on each location outputs $13\times(c+7)$ values.\par
	
	\begin{figure}[!ht]
		\centering
		\includegraphics[width=0.48\textwidth]{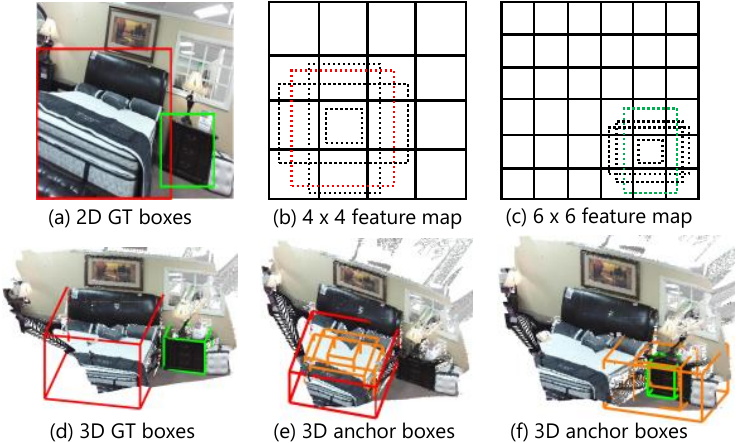}
		\caption{\textbf{An example for finding positive examples.} There are two target objects in the input image, a bed and a night stand. The ground truth of them are shown in (a) and (d) respectively for 2D and 3D. (b) and (c) are two exemplar feature maps used in our model. The 4 dash-line rectangles respectively in (b) and (c) are default boxes related to the location in $3^{rd}$ row and $2^{nd}$ column of (b) and the location in $5^{th}$ row and $5^{th}$ column of (c). During training time, we firstly match default boxes to 2D Ground truth boxes. For example, we have matched red default box with the bed in (b) and green default box with the night stand in (c), which means the location $(3,2)$ in (b) and the location $(5,5)$ in (c) should generate positive 3D anchor boxes. (e) and (f) show 4 anchor boxes produced by the location $(3,2)$ in (b) and the location $(5,5)$ in (c), the sizes of the red 3D box and the green 3D box are mostly close to the bed and night stand, so these two boxes are treated as positives and the rest are negatives. }
		\label{match}
	\end{figure} 
	
	\subsection{Training}
	\noindent\textbf{Positive samples matching:} During training time, we need to identify the positive 3D anchor boxes which are mostly matched with 3D ground truth boxes and train the network accordingly. Consider that $z_{med}$ extracted as the center of anchor boxes is possibly inaccurate due to the noisy depth image, we present a 2D-aided manner for searching of 3D positive samples in this part, which relies on the 2D ground truth box and 2D default boxes defined in SSD\cite{Wei2016SSD}(see Figure \ref{match}).\par
	SSD \cite{Wei2016SSD} has associated a set of default boxes (rectangles with different aspect ratios) with each location of the prediction feature maps and regresses the bounding box offset relative to each default box for 2D object detection. While in our work, default boxes are utilized to determine the locations of the feature maps that are expected to output 3D positive examples. Specifically, if there is at least one default box that is matched with a 2D ground truth box, we consider 3D positive examples are supposed to exist in the 13 anchor boxes attached to that location. Since for each target object the map between 2D ground truth and its 3D ground truth is unique, we can  get the matched 3D ground truth correspondingly. \par
	As the positions of anchor boxes are the same and less accurate, the exact positive anchor box is more dependent on the similarity in size with the ground truth.
	We thus align the centers of these 13 anchor boxes to the matched 3D ground truth box, and choose the one that has the best 3D IoU overlap with the ground truth as positive example, along with the condition that the overlap is higher than a manually defined threshold( which is 0.72 in our experiment). The centers of 3D anchor boxes are shifted only during the matching stage while on regression and evaluation stages they are still used as the original value computed from the depth image. 
	After the matching step, we adopt hard negative mining used in SSD \cite{Wei2016SSD} to choose negative samples.\par
	
	\def \mW {0.033}	
	\begin{table*}[!t]
		{
			\vspace{-1mm}
			\setlength{\tabcolsep}{2.6pt}
			\renewcommand\arraystretch{1.5}
			\centering
			\footnotesize
			\begin{tabular}{l|ccccccccccccccccccc|c|c}
				\hline 
				Methods& \includegraphics[width=\mW\linewidth]{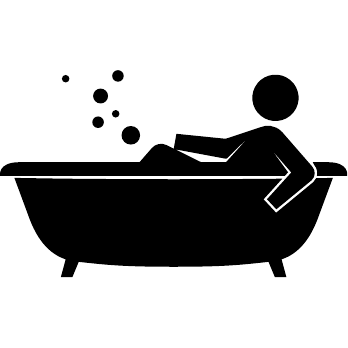} 
				& \includegraphics[width=\mW\linewidth]{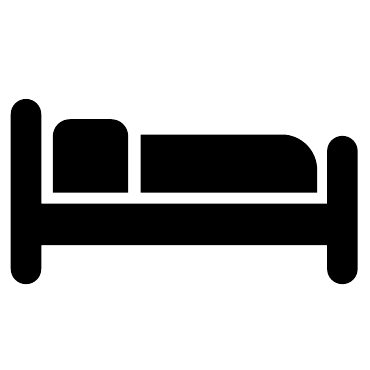} 
				& \includegraphics[width=\mW\linewidth]{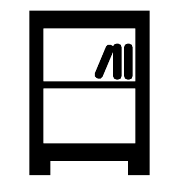} 
				& \includegraphics[width=\mW\linewidth]{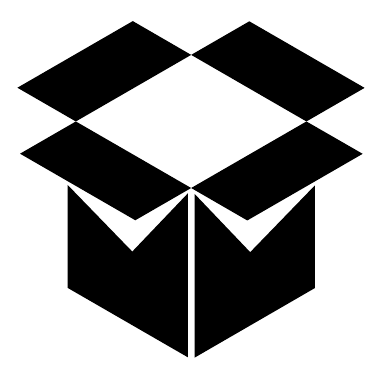} 
				& \includegraphics[width=\mW\linewidth]{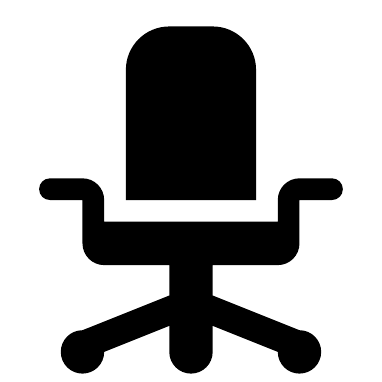}  
				& \includegraphics[width=\mW\linewidth]{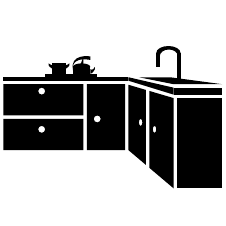}  
				& \includegraphics[width=\mW\linewidth]{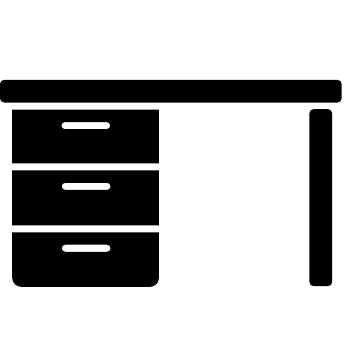}  
				& \includegraphics[width=\mW\linewidth]{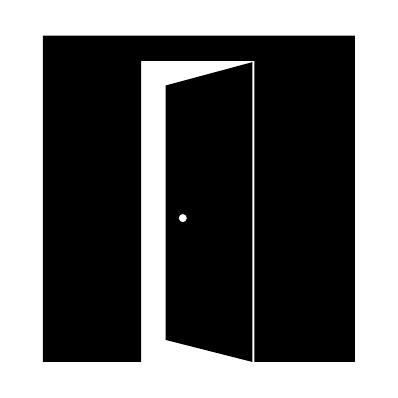}  
				& \includegraphics[width=\mW\linewidth]{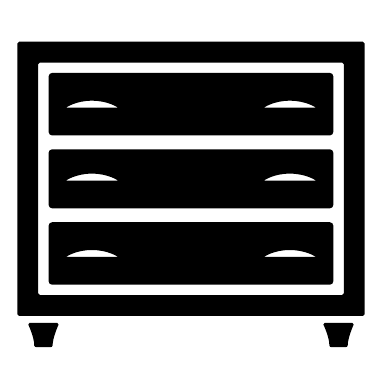}  
				& \includegraphics[width=\mW\linewidth]{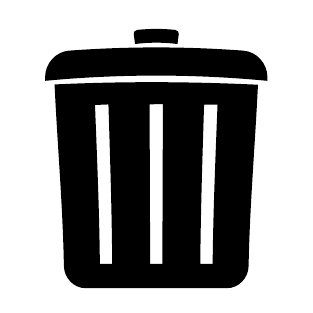}  
				& \includegraphics[width=\mW\linewidth]{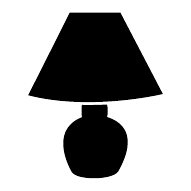}  
				& \includegraphics[width=\mW\linewidth]{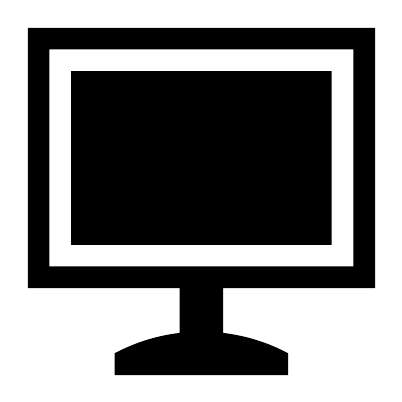}
				& \includegraphics[width=\mW\linewidth]{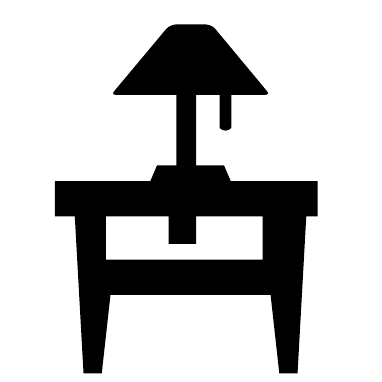}  
				& \includegraphics[width=\mW\linewidth]{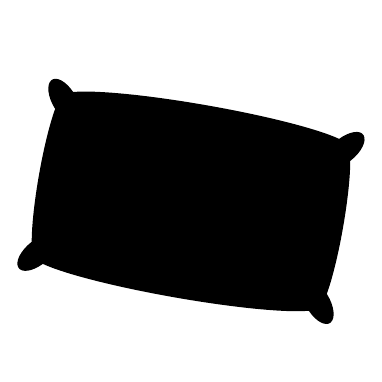}  
				& \includegraphics[width=\mW\linewidth]{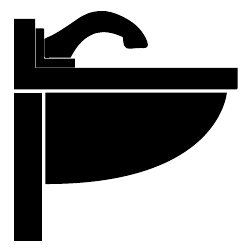}  
				& \includegraphics[width=\mW\linewidth]{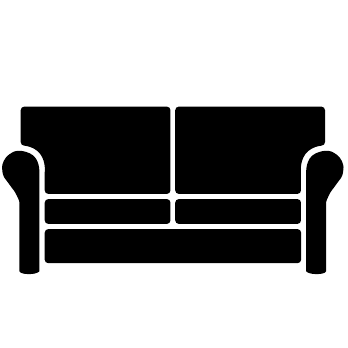}  
				& \includegraphics[width=\mW\linewidth]{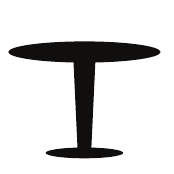}  
				& \includegraphics[width=\mW\linewidth]{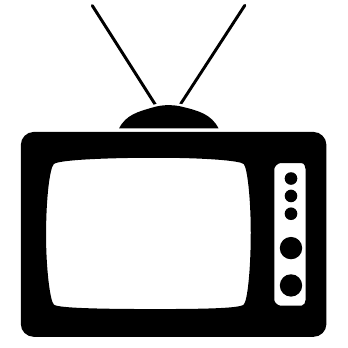} 
				& \includegraphics[width=\mW\linewidth]{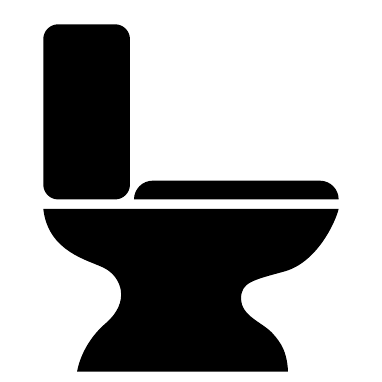}  
				
				& mAP & Runtime \tabularnewline
				\hline
				DSS \cite{Suran2016DSS} & 44.2 & 78.8 & 11.9 & 1.5 & 61.2 & 4.1 & 20.5 & 0.0 & 6.4 & 20.4 & 18.4 & 0.2 & 15.4 & 13.3 & 32.3 & 53.5 & 50.3 & 0.5 & 78.9 & 26.9 & 19.55s \tabularnewline
				\hline
				Ours & 57.1 & 76.2 & 29.4 & 9.2 & 56.8 & 12.3 & 21.9 & 1.7 & 32.5 & 38 & 23.4 & 12.9 & 51.8 & 26.6 & 52.9 & 54.8 & 40.6 & 20.9 & 85.8 & 37.1 & 0.22s \tabularnewline
				\hline 
			\end{tabular}
		}
		\vspace{-1mm}
		\caption{\textbf{Evaluation for 19-class 3D object detection on SUN RGB-D test set.}}
		\label{DSSSUN}
		\vspace{-1mm}  
	\end{table*}
	
	\begin{table*}[t]
		{
			\vspace{-1mm}
			\setlength{\tabcolsep}{8.8pt}
			\renewcommand\arraystretch{1.3}
			\centering
			\footnotesize
			\begin{tabular}{l|cccccccccc|c|c}
				\hline
				Methods 
				&
				\includegraphics[width=0.033\linewidth]{figures/bathtub.pdf} 
				&
				\includegraphics[width=0.033\linewidth]{figures/bed.pdf} 
				&
				\includegraphics[width=0.033\linewidth]{figures/bookshelf.pdf} 
				&
				\includegraphics[width=0.033\linewidth]{figures/chair.pdf} 
				&
				\includegraphics[width=0.033\linewidth]{figures/desk.pdf} 
				&
				\includegraphics[width=0.033\linewidth]{figures/dresser.pdf} 
				&
				\includegraphics[width=0.033\linewidth]{figures/night_stand.pdf} 
				&
				\includegraphics[width=0.033\linewidth]{figures/sofa.pdf} 
				&
				\includegraphics[width=0.033\linewidth]{figures/table.pdf}
				&
				\includegraphics[width=0.033\linewidth]{figures/toilet.pdf}
				& mAP & Runtime \tabularnewline
				\hline
				COG \cite{Zhile2016COG} & 58.26 & 63.67 & 31.8 & 62.17 & 45.19 & 15.47 & 27.36 & 51.02 & 51.29 & 70.07 & 47.63 & 10-30min \tabularnewline
				\hline
				2DD \cite{Jeab20172DD} & 43.45 & 64.48 & 31.40 & 48.27 & 27.93 & 25.92 & 41.92 & 50.39 & 37.02 & 80.4 & 45.1 & 4.15s \tabularnewline
				\hline
				Ours & 57.1 & 76.2 & 29.4 & 56.8 & 21.9 & 32.5 & 51.8 & 54.8 & 40.6 & 85.8&  50.7 & 0.22s \tabularnewline
				\hline
			\end{tabular}
		}
		\vspace{1mm}  
		\caption{\textbf{Evaluation for 10-class 3D object detection on SUN RGB-D test set.}}
		\label{COGSUN}
		\vspace{-3mm}  
	\end{table*}
	\vspace{1mm}
	\noindent\textbf{Multi-task Loss:} Our training loss is similar with that of SSD \cite{Wei2016SSD} however we generalize the formulation to 3D object detection. Assume we have found $N$ positive anchor boxes on the stage of positive samples matching, let $x_{ij}^c=\left\{0,1\right\}$ indicate whether the $i^{th}$ 3D anchor box is matched with the $j^{th}$ 3D ground truth of object class $c$. Our objective loss function is a sum of the 3D bounding box regression loss and the classification loss, denoted as $L_{reg}$ and　$L_{cls}$ respectively:
	\vspace{-1mm}
	\begin{equation}
	L(x,p,r,g) = \frac{1}{N}(L_{cls}(x,p)+L_{reg}(x,r,g))
	\end{equation}
	The loss is set to $0$ when $N$ equal to $0$. $L_{reg}$ is a Smooth L1 loss\cite{R2015Fastrcnn} between the predicted 3D box$(r)$ and the 3D ground truth box$(g)$. We regress offsets for the set $M=\left\{x_{0},y_{0},z_{0},w,h,l,\theta\right\}$ which consists the center$(x_{0},y_{0},z_{0})$ of the 3D anchor box$(d)$, along with its width$(w)$, length$(l)$, height$(h)$ and angle$(\theta)$.
	\vspace{-3mm}
	
	\begin{gather}
	\begin{split}
	L_{reg}(x,r,g) &=\sum_{i\in Pos}^{N}\sum_{m\in M} x_{ij}^{k} smooth_{L1}(r_{i}^{m} - \hat{g}_{j}^{m}) \\
	\hat{g}_{j}^{x_{0}} &=(g_{j}^{x_{0}} - d_{i}^{x_{0}})/d_{i}^{w} \qquad \hat{g}_{j}^{w} =\ln(\frac{g_{j}^{w}}{d_{i}^{w}})\\
	\hat{g}_{j}^{y_{0}} &=(g_{j}^{y_{0}} - d_{i}^{y_{0}})/d_{i}^{l} \qquad \hat{g}_{j}^{l} =\ln(\frac{g_{j}^{l}}{d_{i}^{l }})\\
	\hat{g}_{j}^{z_{0}} &=(g_{j}^{z_{0}} - d_{i}^{z_{0}})/d_{i}^{h} \qquad \hat{g}_{j}^{h} =\ln(\frac{g_{j}^{h}}{d_{i}^{h}})\\
	\hat{g}_{j}^{\theta} &=g_{j}^{\theta} \qquad
	\end{split}
	\end{gather}
	$L_{cls}$ is the classification loss over predicted probabilities for multiple object categories$(p)$ and negative background is labeled as $0$.
	\vspace{-3mm}
	\begin{equation}
	L_{cls}(x,p)=-\sum_{i\in Pos}^{N} x_{ij}^{c} \ln(p_{i}^{c}) - \sum_{i\in Neg} \ln(p_{i}^{0})
	\end{equation} 	
	\noindent\textbf{Training initialization:}\ We initialize our network in the same way on each dataset we used. Firstly, two SSD models are trained for 2D object detection by using 2D bounding box annotations in RGB and depth images, and we get a SSD-RGB model for input RGB images and a SSD-depth model for input depth images. Then we respectively train two base models for 3D object detection by using only one form of image data (RGB or depth images). The base model is composed of the VGG-16 network in the \emph{hierarchical feature fusion} part and the upper 8 auxiliary convolution layers in the \emph{multi-layer prediction} part. When training the model for RGB images,  we have relied on the depth image to estimate the 3D positions for 3D anchor boxes. Thus we get a 3D-RGB model for RGB images and a 3D-depth model for depth images. The parameters of the VGG-16 network in the 3D-RGB model are initialized by weights from the SSD-RGB model and the parameters of the  VGG-16 network in the 3D-depth model are initialized by weights from the SSD-depth model.\par
	Finally, when training our 3D-SSD, the parameters of two VGG-16 networks in the \emph{hierarchical feature fusion} part are separately initialized by weights from the 3D-RGB and 3D-depth models. \par
	\vspace{1mm}
	\noindent\textbf{Optimization:} Our 3D-SSD network is trained with stochastic gradient descent(SGD). The network on the SUN training set is trained with a learning rate of $2.5\times10^{-4}$ for 60K iterations. Then we reduce the learning rate to $2.5\times10^{-5}$ for another 20K iterations. The network on the NYUv2 training set is trained for 27200 iterations with an initial learning rate of $2.5\times10^{-4}$, which is reduced to $2.5\times10^{-5}$at 25600 iterations. We employ the data augmentation similar to SSD\cite{Wei2016SSD} and add horizontally flipped images to the training set. No other extra data is used during training. When training the network on the NYUv2 training set, we do not use data on the SUN training set.  
	
	\section{Experiments}
	\begin{table*}[t]
		{
			\vspace{-1mm}
			\setlength{\tabcolsep}{2.6pt}
			\renewcommand\arraystretch{1.5}
			\centering
			\footnotesize
			\begin{tabular}{l|ccccccccccccccccccc|c|c}
				\hline 
				Methods& \includegraphics[width=\mW\linewidth]{figures/bathtub.pdf} 
				& \includegraphics[width=\mW\linewidth]{figures/bed.pdf} 
				& \includegraphics[width=\mW\linewidth]{figures/bookshelf.pdf} 
				& \includegraphics[width=\mW\linewidth]{figures/box.pdf} 
				& \includegraphics[width=\mW\linewidth]{figures/chair.pdf}  
				& \includegraphics[width=\mW\linewidth]{figures/counter.pdf}  
				& \includegraphics[width=\mW\linewidth]{figures/desk.pdf}  
				& \includegraphics[width=\mW\linewidth]{figures/door.pdf}  
				& \includegraphics[width=\mW\linewidth]{figures/dresser.pdf}  
				& \includegraphics[width=\mW\linewidth]{figures/garbage_bin.pdf}  
				& \includegraphics[width=\mW\linewidth]{figures/lamp.pdf}  
				& \includegraphics[width=\mW\linewidth]{figures/monitor.pdf}
				& \includegraphics[width=\mW\linewidth]{figures/night_stand.pdf}  
				& \includegraphics[width=\mW\linewidth]{figures/pillow.pdf}  
				& \includegraphics[width=\mW\linewidth]{figures/sink.pdf}  
				& \includegraphics[width=\mW\linewidth]{figures/sofa.pdf}  
				& \includegraphics[width=\mW\linewidth]{figures/table.pdf}  
				& \includegraphics[width=\mW\linewidth]{figures/tv.pdf} 
				& \includegraphics[width=\mW\linewidth]{figures/toilet.pdf}
				& mAP & Runtime\tabularnewline
				\hline
				DSS \cite{Deng2017} & 62.3 & 81.2 & 23.9 & 3.8 & 58.2 & 24.5 & 36.1 & 0.0 & 31.6 & 27.2 & 28.7 & 2.0 & 54.5 & 38.5 & 40.5 & 55.2 & 43.7 & 1.0 & 76.3 & 36.3 & 19.55s\tabularnewline
				\hline
				Deng \cite{Deng2017} & 36.1 &84.5 &40.6 &4.9 &46.4 &44.8& 33.1& 10.2& 44.9& 33.3 &29.4& 3.6 &60.6 &46.3& 58.3& 61.8 &43.2& 16.3& 79.7 &40.9 &85s \tabularnewline
				\hline
				Ours & 48.9 & 84 & 26.1 & 2.2 & 50.7 & 44.4 & 32.8 & 9.2 & 29.1 & 30.8 & 32.2 & 11.2 & 64.1 & 40.2 & 64.1 & 57.8 & 39 & 9.1 & 79.4 & 39.7 & 0.22s \tabularnewline
				\hline
			\end{tabular}
		}
		\vspace{-1mm} 
		\caption{\textbf{Evaluation for 19-class 3D object detection on NYUv2 RGB-D test set.}}
		\label{Deng}
		\vspace{-2mm}
	\end{table*}
	
	\begin{table*}[t]
		{
			\setlength{\tabcolsep}{2.3pt}
			\renewcommand\arraystretch{1.5}
			\centering
			\footnotesize
			\begin{tabular}{l|ccccccccccccccccccc|c|c}
				\hline 
				Data& \includegraphics[width=\mW\linewidth]{figures/bathtub.pdf} 
				& \includegraphics[width=\mW\linewidth]{figures/bed.pdf} 
				& \includegraphics[width=\mW\linewidth]{figures/bookshelf.pdf} 
				& \includegraphics[width=\mW\linewidth]{figures/box.pdf} 
				& \includegraphics[width=\mW\linewidth]{figures/chair.pdf}  
				& \includegraphics[width=\mW\linewidth]{figures/counter.pdf}  
				& \includegraphics[width=\mW\linewidth]{figures/desk.pdf}  
				& \includegraphics[width=\mW\linewidth]{figures/door.pdf}  
				& \includegraphics[width=\mW\linewidth]{figures/dresser.pdf}  
				& \includegraphics[width=\mW\linewidth]{figures/garbage_bin.pdf}  
				& \includegraphics[width=\mW\linewidth]{figures/lamp.pdf}  
				& \includegraphics[width=\mW\linewidth]{figures/monitor.pdf}
				& \includegraphics[width=\mW\linewidth]{figures/night_stand.pdf}  
				& \includegraphics[width=\mW\linewidth]{figures/pillow.pdf}  
				& \includegraphics[width=\mW\linewidth]{figures/sink.pdf}  
				& \includegraphics[width=\mW\linewidth]{figures/sofa.pdf}  
				& \includegraphics[width=\mW\linewidth]{figures/table.pdf}  
				& \includegraphics[width=\mW\linewidth]{figures/tv.pdf} 
				& \includegraphics[width=\mW\linewidth]{figures/toilet.pdf}
				& mAP & Runtime\tabularnewline
				\hline
				RGB & 42.9 & 68.2 & 22.5& 10.4 & 48.6 & 10.6 & 16.4 & 0.4 & 26 & 33.8 & 17.1 & 13.5 & 42 & 16 & 45.3 & 46.2 & 34.6 & 14.8 & 80.1 & 31 & 0.21s\tabularnewline
				\hline
				HHA & 57.1& 74.5 & 18.9 & 3.7 & 55.3 & 6.6 & 17.3 & 0.1 & 26.7 & 33.1 & 17.4 & 13.8 & 48.2 & 22.7 & 48.1 & 52.2 & 37.9 & 15.2 & 78.9 & 33& 0.21s\tabularnewline
				\hline
				Depth & 53.7 & 74.4 & 19.1 & 9.7 & 53.8 & 4.5 & 19.2 & 1.3 & 22.8 & 27.6 & 23 & 8.4 & 47.5 & 22.5 & 40.2 & 51.1 & 38.9 & 7.6 & 79 & 31.8 & 0.21s\tabularnewline
				\hline
				RGB+HHA & 59 & 78 & 26.1 & 7 & 57.6 & 11.7 & 22.4 & 0.3 & 32.8 & 40.4 & 18.3 & 12.9 & 51.2 & 26.8 & 51.1 & 54.8 & 42.1 & 19.9 & 84.6 & 36.7 & 0.22s\tabularnewline
				\hline
				RGB+Depth & 57.1 & 76.2 & 29.4 & 9.2 & 56.8 & 12.3 & 21.9 & 1.7 & 32.5 & 38 & 23.4 & 12.9 & 51.8 & 26.6 & 52.9 & 54.8 & 40.6 & 20.9 & 85.8 & 37.1 & 0.22s\tabularnewline  	
				\hline
			\end{tabular}
		}
		\vspace{-1mm} 
		\caption{\textbf{An ablation study of different features: Performances are evaluated on SUN RGB-D test set.}}
		\label{DSSinSUN}
		\vspace{-3mm}
	\end{table*}
	
	We evaluated our framework on the challenging SUN RGB-D dataset \cite{Suran2015SUN} and  NYUv2 dataset \cite{Silberman2012NYU}. The former dataset consists of 5285 images for training and 5050 images for testing, while the NYUv2 is a bit smaller that contains 795 images for training and 654 images for testing, whose annotations are improved by \cite{Deng2017} for evaluation.\par
	\vspace{1mm}
	\noindent\textbf{Evaluation Metric:} We evaluate the performance of different methods using the 3D volume Intersection over Union (IoU) metric defined in \cite{Surans2014sliding}. A detection is considered as a true positive if its IoU with the ground truth is larger than 0.25. Similar to \cite{Suran2016DSS} and \cite{Deng2017}, we trained our model for 19 object classes detection, and calculated mean Average Precision for an overall accuracy comparison. Moreover, we have made a comparison on the computation efficiency, which is evaluated as the total time from the input of data to the output of 3D bounding boxes.\par
	
	\vspace{1mm}
	\noindent\textbf{Comparisons with the state-of-art works:} Table \ref{DSSSUN} and Table \ref{COGSUN} show the quantitative results of the proposed method with three state-of-the-art works for amodal 3D object detection on the SUN RGB-D datasets. During testing, 3D-SSD takes 0.22s per RGB-D image pair on a  Nvidia Titan X GPU, which is the fastest reported in 3D object detection to our knowledge. Our method outperforms DSS\cite{Suran2016DSS} by a significant margin of 10.2\% mAP and is $88\times$ faster. When compared with COG \cite{Zhile2016COG} and 2DD \cite{Jeab20172DD}, we choose the results of the same 10 object classes as they reported and recomputed mAP for them. The accuracy mAP is 3\% higher than \cite{Zhile2016COG} and 5.6\% higher than \cite{Jeab20172DD}. Moreover, we have significantly improved the speed which is 18$\times$ faster than the current level reported (\cite{Jeab20172DD}). Different from DSS, COG and 2DD that all align detected object with the estimated room layout, our method predicts orientations without Manhattan world assumption and thus performs better for indoor scenes where objects poses vary greatly. \par
	
	DSS, 2DD and COG have all extracted features of point clouds that are restructured from depth images which is usually sparse and noisy, especially for small objects (\eg tv, monitor, lamp, garbage bin, pillow) and partially-visible objects (\eg dresser), which leads to a less distinctive feature. In contrast, our approach explores dense and contiguity characteristics in the initial RGB and depth images to avoid the impact of the sparsity of 3D data and therefore is more robust to the aforementioned object categories.\par
	
	2DD uses single RGB images to detect objects on the image plane, and then regresses 3D positions and sizes of each detected 2D bounding box by using the 3D information from the depth image. This separate processing on RGB and depth images is incapable to integrate complementary information of texture and geometric features, and resulting in a less satisfactory performance. \par
	
	Moreover, we tested our algorithm on the NYUv2 dataset to compare with another state-of-the-art method of Deng\cite{Deng2017}. The input images are resized to 300$\times$300 in our network, whereas Deng uses the original image sizes (561$\times$427). Results in Table \ref{Deng} shows our approach achieves comparable performance to their method with a relatively smaller input image sizes, and runs at a much faster speed. \par
	
\begin{figure*}[t]
		\vspace{-1mm} 
		\includegraphics[width=1\textwidth]{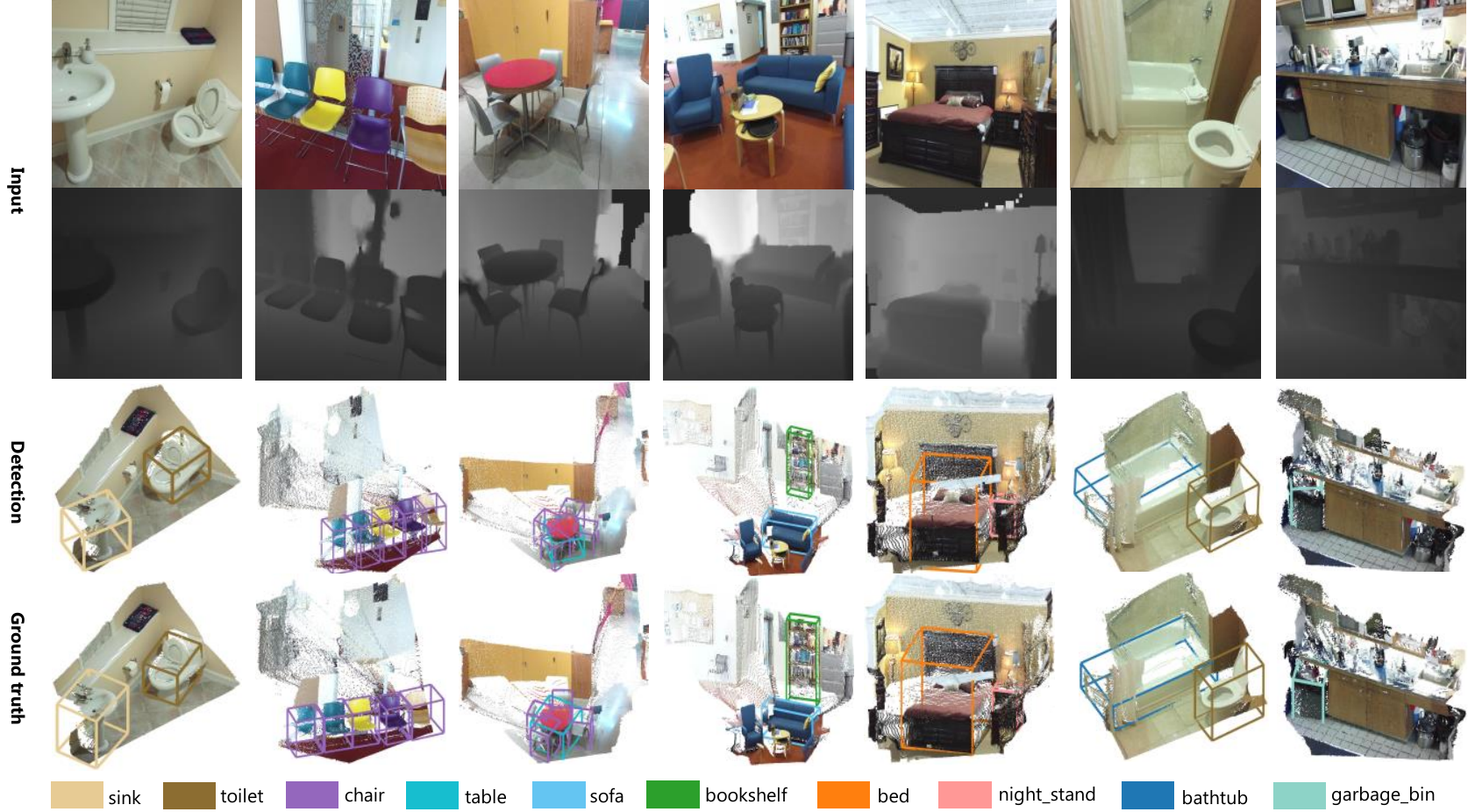}
		\caption{{\small Examples of detected objects in SUN RGB-D datasets. We show detections with scores higher than a threshold$(0.55)$. Each color corresponds to an object category. The $1^{st}$ row  and $2^{nd}$ row show the input RGB and depth images. Detected boxes are shown in the $3^{rd}$ row. Ground truth boxes are shown in the $4^{th}$ row.}}
		\label{results}
	\end{figure*}

\begin{figure*}[t]
		\vspace{-1mm} 
		\centering
		\includegraphics[width=1\textwidth]{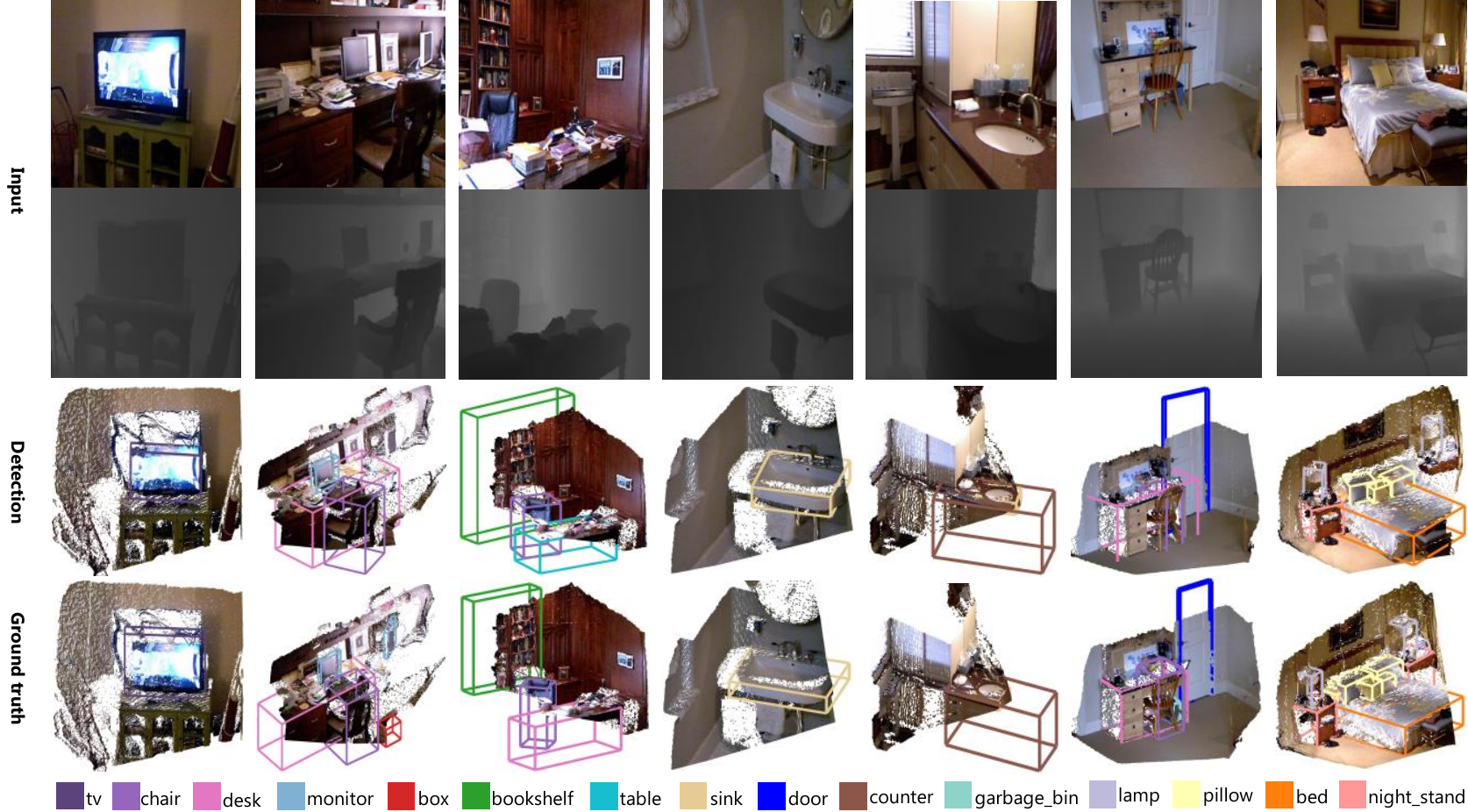}
		\caption{{\small Examples of detected objects in NYUV2 datasets. We show detections with scores higher than a threshold$(0.55)$. Each color corresponds to an object category. The $1^{st}$ row  and $2^{nd}$ row show the input RGB and depth images. Detected boxes are shown in the $3^{rd}$ row. Ground truth boxes are shown in the $4^{th}$ row.}}
		\label{results-nyu}
	\end{figure*}

	Some qualitative results on the SUN RGB-D data and the NYUv2 data are shown in Figure \ref{results} and Figure \ref{results-nyu} respectively. Detection results indicate our 3D-SSD can estimate the full extent of objects in 3D space even under partial observation(\eg truncated sink, chair, toilet and occluded chair, bathtub in Figure \ref{results}; truncated desk, chair, counter, bed and occluded chair in Figure \ref{results-nyu}). 
	
	Then, we will present the results of some control experiments conducted to analyze the contributions of each component in our model.\par
	\begin{table}[ht]
			\setlength{\tabcolsep}{3.1pt}
			\renewcommand\arraystretch{1.2}
			{
				\centering
				\footnotesize
				\vspace{2mm}
				\begin{tabular}{cccccc|c}
					\hline
					\multicolumn{6}{c|}{Hierarchically fused layers from:}
					& \tabularnewline  
					conv4-3s & conv7s & conv8-2s & conv9-2s & conv10-2s & conv11-2s & mAP \tabularnewline
					\hline
					\Checkmark & & & & & & 27.6 \tabularnewline
					\Checkmark & \Checkmark & & & & & 37.1\tabularnewline
					\Checkmark & \Checkmark & \Checkmark & & & & 36.8\tabularnewline
					\Checkmark & \Checkmark & \Checkmark & \checkmark & & & 36.8\tabularnewline
					\Checkmark & \Checkmark & \Checkmark & \checkmark & \checkmark & & 36.6\tabularnewline
					\Checkmark & \Checkmark & \Checkmark & \checkmark & \checkmark & \checkmark & 36.0\tabularnewline  	
					\hline
				\end{tabular}
			}
			\vspace{-1mm} 
			\caption{\textbf{Effects of different feature concatenation strategies.} Performance are evaluated on SUN RGB-D test set.}
			\label{hierarchical}
			\vspace{-5mm}
		\end{table}
	
	\vspace{1mm}
	\noindent\textbf{Multi-modal features:} To study the effectiveness of the conjunction of features from different input images, we performed experiments with different combination styles of RGB image, depth image and the HHA image \cite{S2014rich}(encodes depth image as Horizontal disparity, Height above ground, and angle of local surface normals within estimated gravity direction). The results are shown in the Table \ref{DSSinSUN}. When only using one form of image data as input, RGB, depth and HHA perform almost comparably(the training for RGB images also involve the depth images to estimate the 3D position). Combining the RGB and depth (or HHA) images outperforms individual RGB image inputs, which leads to the conclusion that the incorporation of features learned from the depth image to that learned from the RGB image can contribute to the improving of detection result, mainly due to the additional geometric information that is not available in the RGB images. Further, the Table \ref{DSSinSUN} shows the depth information is better at detecting objects with less clear textures (\eg bathtub, bed, chair, night stand, pillow and table). It also indicates the objects with high texture can achieve better performance when adding in the depth information(\eg bookshelf, counter, dresser, garbage bin, lamp, bathtub). Therefore, the algorithm is able to synergistically fuse information from multi-modal information. \par
	\vspace{1mm}
	\noindent\textbf{Hierarchical fusion:} To illustrate the improvement on accuracy from the proposed hierarchical fusion structure, we conducted another comparison experiment with different hierarchical level that ranges from one-stage fusion (one pair of layers) to six-stage fusion (six pairs of layers), as shown in Table \ref{hierarchical}. For the one-stage fusion, we concatenated the convolution layers of conv4-3-rgb and conv4-3-d in the \emph{hierarchical feature fusion} part, followed by two $1\times1$ convolution layers. Then the fused feature layer directly goes through the following 5 convolution layers of the VGG-16 network and the upper 8 convolution layers in the \emph{multi-layer prediction} part. Thus the feature layers conv4-5, conv7, conv8-2, conv9-2, conv10-2, conv11-2 are used for predicting detections. Then we gradually increase the hierarchical fusion level by adding more layers to concatenate features one by one and compare their results. Not surprisingly, fusion in one stage performs worst and is less superior than the two-stage fusion by 9.5\% in mAP. The accuracy has reached to a saturation after the fusion level of two-stage. Further increasing the hierarchical level has no obvious improvement however make the network more complicated, and thus we have finally choose the two-stage fusion strategy in our 3D-SSD model. \par
	\vspace{-5mm} 
	\section{Conclusion}
	\vspace{-4mm} 
	We have proposed a novel neural network for amodal 3D object detection in indoor scenes. Our model was designed to exploit the complementary information in RGB and depth images. A hierarchical fusion structure was presented to concatenate features from different input data and multiple feature layers were considered to carry out 3D bounding box regression along with object classification. The structure managed to capture fine details as well as contextual information of the scene images. Experiments on publicly available datasets showed our method significantly outperformed the state-of-the-art methods in terms of both accuracy and computation efficiency for 3D object detection.\par
	\vspace{15mm} 
	
	{\small
		\bibliographystyle{ieee}
		\bibliography{egbib}

\begin{thebibliography}{10}\itemsep=-1pt

\bibitem{P2014MCG}
P.~Arbelaez, J.~Pont-Tuset, J.~T. Barron, F.~Marques, and J.~Malik.
\newblock Multiscale combinatorial grouping.
\newblock In {\em CVPR}. 2014.

\bibitem{Cai2016early}
Z.~Cai, Q.~Fan, R.~Feris, and N.~Vasconcelos.
\newblock A unified multi-scale deep convolutional neural network for fast
  object detection.
\newblock In {\em ECCV}. 2016.

\bibitem{CPMC2012}
J.~Carreira and C.~Sminchisescu.
\newblock Cpmc: Automatic object segmentation using constrained parametric
  min-cuts.
\newblock In {\em TPAMI}. 2012.

\bibitem{Multiview2017}
X.~Chen, H.~Ma, J.~Wan, B.~Li, and T.~Xia.
\newblock Multi-view 3d object detection network for autonomous driving.
\newblock In {\em CVPR}. 2017.

\bibitem{Deng2017}
Z.~Deng and L.~J. Latecki.
\newblock Amodal detection of 3d objects: Inferring 3d bounding boxed from 2d
  ones in rgb-depth images.
\newblock In {\em CVPR}. 2017.

\bibitem{VOC2010}
M.~Everingham, L.~V. Gool, C.~K. Williams, J.~Winn, and A.~Zisserman.
\newblock The pascal visual object classes (voc) challenge.
\newblock In {\em IJCV}. 2010.

\bibitem{DPM2010}
P.~Felzenszwalb, R.~Girshick, D.~McAllester, and D.~Ramanan.
\newblock Object detection with discriminatively trained part-based models.
\newblock In {\em TPAMI}. 2010.

\bibitem{R2015Fastrcnn}
R.~Girshick.
\newblock Fast r-cnn.
\newblock In {\em ICCV}. 2015.

\bibitem{R2014rcnn}
R.~Girshick, J.~Donahue, T.~Darrell, and J.~Malik.
\newblock Rich feature hierarchies for accurate object detection and
  segmentation.
\newblock In {\em CVPR}. 2014.

\bibitem{Saurabh2015depthrcnn}
S.~Gupta, P.~Arbelaez, R.~Girshick, and J.~Malik.
\newblock Aligning 3d models to rgb-d images of cluttered scenes.
\newblock In {\em CVPR}. 2015.

\bibitem{S2014rich}
S.~Gupta, R.~Girshick, P.~Arbelaez, and J.~Malik.
\newblock Learning rich features from rgb-d images for object detection and
  segmentation.
\newblock In {\em ECCV}. 2014.

\bibitem{He2017maskrcnn}
K.~He, G.~Gkioxari, P.~Dollár, and R.~Girshick.
\newblock Mask r-cnn.
\newblock In {\em ICCV}. 2017.

\bibitem{Jeab20172DD}
J.~Lahoud and B.~Ghanem.
\newblock 2d-driven 3d object detection in rgb-d images.
\newblock In {\em ICCV}. 2017.

\bibitem{T2017FCN}
T.-Y. Lin, P.~Dollar, R.~Girshick, K.~He, B.~Hariharan, and S.~Belongie.
\newblock Feature pyramid networks for object detection.
\newblock In {\em CVPR}. 2017.

\bibitem{He2017focal}
T.-Y. Lin, P.~Goyal, R.~Girshick, K.~He, and P.~Dollár.
\newblock Focal loss for dense object detection.
\newblock In {\em ICCV}. 2017.

\bibitem{COCO2014}
T.-Y. Lin, M.~Maire, S.~Belongie, J.~Hays, P.~Perona, D.~Ramanan, P.~Dolĺar,
  and C.~L. Zitnick.
\newblock Microsoft coco: Com- mon objects in context.
\newblock In {\em ECCV}. 2014.

\bibitem{Wei2016SSD}
W.~Liu, D.~Anguelov, D.~Erhan, C.~Szegedy, and S.~Reed.
\newblock Ssd: Single shot multibox detector.
\newblock In {\em ECCV}. 2016.

\bibitem{Wei2015DBM}
W.~Liu, R.~Ji, and S.~Li.
\newblock Towards 3d object detection with bimodal deep boltzmann machines over
  rgbd imagery.
\newblock In {\em CVPR}. 2015.

\bibitem{Silberman2012NYU}
S.~Nathan, H.~Derek, K.~Pushmeet, and R.~Fergus.
\newblock Indoor segmentation and support inference from rgbd images.
\newblock In {\em ECCV}. 2012.

\bibitem{J2016YOLO}
J.~Redmon, S.~Divvala, R.~Girshick, and A.~Farhadi.
\newblock You only look once: Unified, real-time object detection.
\newblock In {\em CVPR}. 2016.

\bibitem{J2017YOLO9000}
J.~Redmon and A.~Farhadi.
\newblock Yolo9000: Better, faster, stronger.
\newblock In {\em CVPR}. 2017.

\bibitem{He2014fasterrcnn}
S.~Ren, K.~He, R.~Girshick, and J.~Sun.
\newblock Faster r-cnn: Towards real-time object detection with region proposal
  networks.
\newblock In {\em NIPS}. 2015.

\bibitem{Zhile2016COG}
Z.~Ren and E.~B. Sudderth.
\newblock Three-dimensional object detection and layout prediction using clouds
  of oriented gradients.
\newblock In {\em CVPR}. 2016.

\bibitem{imagenet2015}
O.~Russakovsky, J.~Deng, H.~Su, J.~Krause, S.~Satheesh, S.~Ma, Z.~Huang,
  A.~Karpathy, A.~Khosla, M.~Bernstein, and et~al.
\newblock Imagenet large scale visual recognition challenge.
\newblock In {\em IJCV}. 2015.

\bibitem{Suran2015SUN}
S.~Song, S.~P. Lichtenberg, and J.~Xiao.
\newblock Sun rgb-d: A rgb-d scene understanding benchmark suite.
\newblock In {\em ECCV}. 2015.

\bibitem{Surans2014sliding}
S.~Song and J.~Xiao.
\newblock Sliding shapes for 3d object detection in depth images.
\newblock In {\em ECCV}. 2014.

\bibitem{Suran2016DSS}
S.~Song and J.~Xiao.
\newblock Deep sliding shapes for amodal 3d object detection in rgb-d images.
\newblock In {\em CVPR}. 2016.

\bibitem{K2013accurate}
B.~soo Kim, S.~Xu, and S.~Savarese.
\newblock Accurate localization of 3d object from rgb-d data using segmentation
  hypothesis.
\newblock In {\em ICCV}. 2013.

\bibitem{D2013Holistic}
B.~soo Kim, S.~Xu, and S.~Savarese.
\newblock Holistic scene understanding for 3d object detection with rgbd
  cameras.
\newblock In {\em ICCV}. 2013.

\end{thebibliography}


\begin{thebibliography}{1}\itemsep=-1pt

\bibitem{Suran2015SUN}
S.~P.~L. Shuran~Song and J.~Xiao.
\newblock Sun rgb-d: A rgb-d scene understanding benchmark suite.
\newblock In {\em ECCV}, pages 567--576. Springer, 2015.

\end{thebibliography}
	}
	
\end{document}